\documentclass[runningheads]{llncs}

\usepackage{eccv}

\usepackage{eccvabbrv}

\usepackage{graphicx}
\usepackage{booktabs}
\usepackage{multirow}
\usepackage{makecell}
\usepackage{caption}
\usepackage{colortbl}
\usepackage[table]{xcolor}
\definecolor{greenhighlight}{RGB}{235, 250, 235}
\definecolor{headergray}{gray}{0.92}

\usepackage{soul}

\newcommand\blfootnote[1]{%
  \begingroup
  \renewcommand\thefootnote{}\footnote{#1}%
  \addtocounter{footnote}{-1}%
  \endgroup
}

\usepackage{xspace}
\newcommand{\method}{SCORE\xspace}
\newcommand{\res}[2]{#1_{\pm \mathit{#2}}}

\usepackage[accsupp]{axessibility}  

\usepackage[pagebackref,breaklinks,colorlinks,citecolor=eccvblue]{hyperref}

\usepackage{orcidlink}

\usepackage[most]{tcolorbox} 
\newtcolorbox{promptbox}[1]{
  enhanced,
  fonttitle=\large\bfseries\ttfamily, 
  title={#1},
  colframe=gray,
  colbacktitle=white, 
  coltitle=black,
  colback=gray!5,
  fontupper=\ttfamily\small,
  titlerule=0.5pt, titlerule style=gray,
  boxrule=0.5pt, arc=0mm
}

\begin{document}

\title{Dynamic Token Compression for Efficient Video Understanding through Reinforcement Learning} 

\titlerunning{SCORE}

\author{Shida Wang\textsuperscript{$*$}\inst{1,2} \and
YongXiang Hua\textsuperscript{$*$}\inst{1,2} \and
Zhou Tao\inst{1,2} \and
Haoyu Cao\inst{1,2} \and \\
Linli Xu\textsuperscript{$\dagger$}\inst{1,2}
}

\authorrunning{S. Wang \and Y. Hua et al.}

\institute{University of Science and Technology of China \and
State Key Laboratory of Cognitive Intelligence\\
\email{\{wangshida, yx15333063290, zhoutao24, caohaoyu\}@mail.ustc.edu.cn}\\
\email{linlixu@ustc.edu.cn}}

\maketitle

\blfootnote{$*$ Equal contribution.\ \ $\dagger$ Corresponding author. }

\begin{abstract}

Multimodal Large Language Models have demonstrated remarkable capabilities in video understanding, yet face prohibitive computational costs and performance degradation from ``context rot'' due to massive visual token redundancy.
Existing compression strategies typically rely on heuristics or fixed transformations that are often decoupled from the downstream task objectives, limiting their adaptability and effectiveness.
To address this, we propose \textbf{\method} (\textbf{S}urprise-augmented token \textbf{CO}mpression via \textbf{RE}inforcement learning), a unified framework that learns an adaptive token compression policy.
SCORE introduces a lightweight policy network conditioned on a surprise-augmented state representation that incorporates inter-frame residuals to explicitly capture temporal dynamics and motion saliency.
We optimize this policy using a group-wise reinforcement learning scheme with a split-advantage estimator, stabilized by a two-stage curriculum transferring from static pseudo-videos to real dynamic videos.
Extensive experiments on diverse video understanding benchmarks demonstrate that \method significantly outperforms state-of-the-art baselines.
Notably, \method achieves a \textbf{16$\times$} prefill speedup while preserving 99.5\% of original performance at a 10\% retention ratio, offering a scalable solution for efficient long-form video understanding.

\keywords{Video compression \and Reinforcement learning \and Multimodal large language models }
\end{abstract}
\begingroup
\begin{figure}[t]
    \centering
    \includegraphics[width=0.8\linewidth]{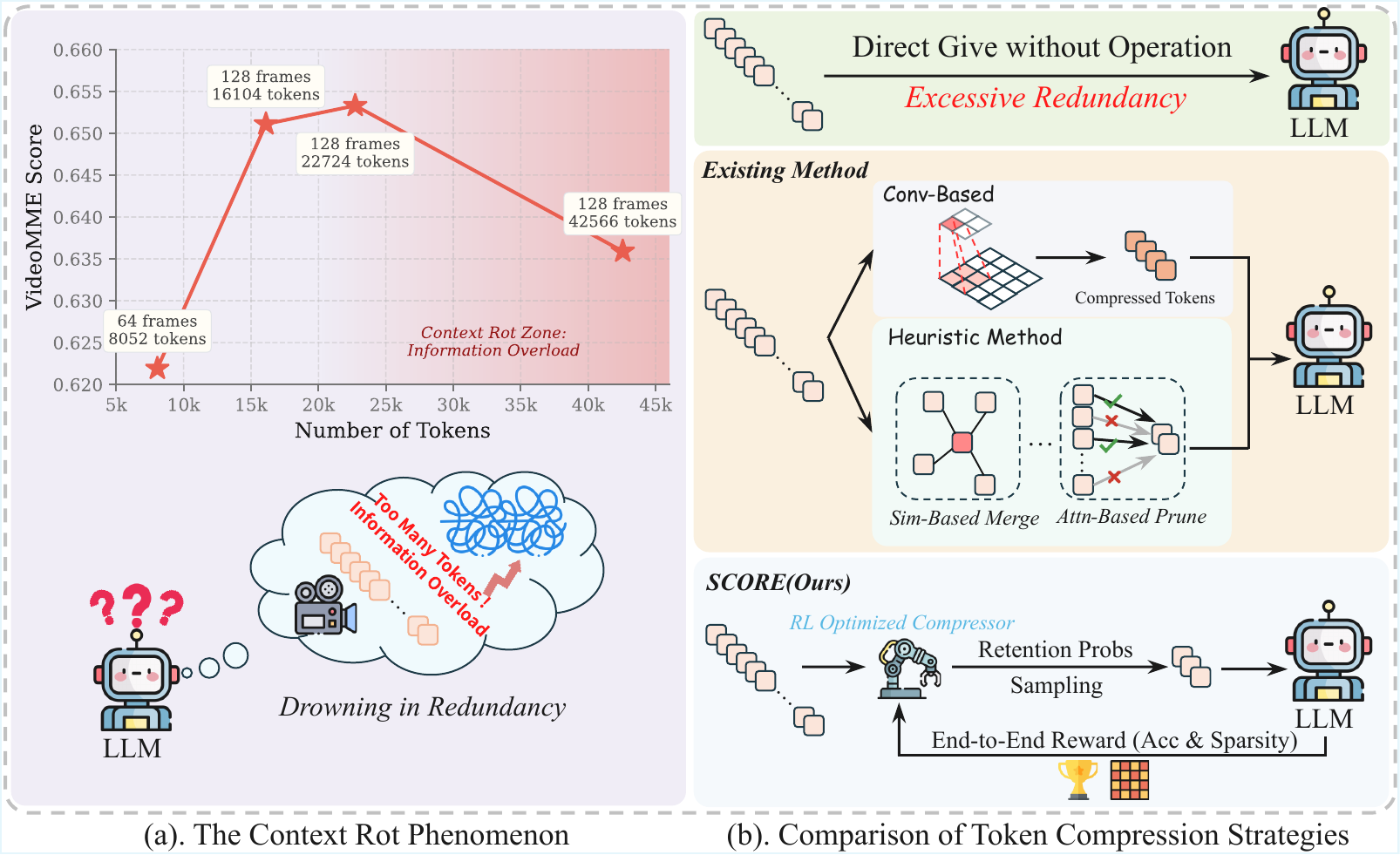}
    \caption{\textbf{(a)} Visual token redundancy induces context rot in video MLLMs. \textbf{(b)} A comparision of existing token compression methods and \method.}
    \label{fig:intro}
\end{figure}
\endgroup

\section{Introduction}\label{sec:intro}

A hallmark of modern Large Language Models (LLMs) is their remarkable capacity to process extensive context windows.
However, as input length increases, these models suffer from \textit{context rot}~\cite{hong2025context}, where their capacity to utilize information diminishes, particularly for tokens in the middle of the sequence.
This degradation leads to notable performance drops on tasks that require holistic reasoning over extended contexts.
The problem of context rot is further exacerbated in video understanding scenarios involving Multimodal Large Language Models (MLLMs)~\cite{bai2025qwen2,bai2025qwen3vltechnicalreport,wang2025internvl3,zhang2024video,team2025kimi}.
Standard vision encoders produce tens of thousands of tokens per video, often dominated by redundant static backgrounds and repetitive actions.
This redundancy imposes prohibitive computational overhead due to the quadratic complexity of self-attention, and exacerbates ``context rot,''  drowning out essential information.
As illustrated in \cref{fig:intro}(a), model performance deteriorates when the token count becomes excessive.
This behavior directly reflects the manifestation of context rot, in which semantically essential information is effectively drowned out by a large volume of irrelevant or redundant tokens.\looseness=-1

This observation underscores the critical need for intelligent \emph{token compression}~\cite{chen2024image,shen2024longvu,fu2024framefusion,xu2025slowfast}, which aims to dynamically filter the visual token stream and retain only the most informative subset before feeding it into the LLM.
As shown in \cref{fig:intro}(b), existing approaches can be broadly categorized into two groups, both of which suffer from notable limitations.
Transformation-based methods~\cite{zhang2024video,xu2024pllava,maaz2024video,cheng2024videollama}, such as convolution or pooling, apply uniform operations across tokens and inherently lack the content adaptivity and semantic discrimination required to preserve critical information while eliminating redundancy. 
Heuristic methods~\cite{tao2025dycoke,shen2025fastvid,shao2025holitom}, including those based on similarity measures or clustering, are often decoupled from the model's forward pass.
This separation introduces an optimization gap between the compression heuristic and the downstream objective, often resulting in suboptimal performance across varying videos and tasks.
Therefore, effectively mitigating context rot calls for a compression strategy that is content-aware, temporally grounded, and jointly optimized with the video understanding model.

To address these challenges, we propose \textbf{\method}, a reinforcement learning framework that learns an adaptive token compression policy in an end-to-end manner.
At its core, \method introduces a lightweight \emph{token-level visual compressor} that is inserted between a frozen vision encoder and the LLM.
To effectively mitigate temporal redundancy, the compressor constructs its input by jointly considering each token's embedding and a \textit{surprise signal} quantifying inter-frame variations, thereby emphasizing regions exhibiting significant temporal changes.
The resulting policy network outputs a retention probability for each visual token, and we optimize this discrete decision policy using a group-based reinforcement learning strategy.
For each video, we sample multiple masks from the policy and evaluate them by running the frozen LLM to obtain generation-quality rewards.
We then apply a split-advantage estimator that promotes higher sparsity only when the reward remains competitive.
Crucially, to cope with the vast combinatorial action space, we further introduce \textit{two-stage curriculum learning}: a pseudo-video warm-up with high-contrast residuals followed by real-video training, which bootstraps learning with less noisy temporal residual cues before adapting to subtle motion in real videos.
This integrated approach, which combines surprise-aware state representations, performance-driven group optimization, and progressive curriculum learning, enables \method to dynamically distill long videos into compact yet informative subsets of visual tokens.
As a result, \method effectively mitigates context rot, substantially reduces computational overhead, and maintains or even improves model performance.

Extensive experiments on diverse video understanding benchmarks demonstrate the effectiveness of \method. 
Across a broad range of retention ratios, \method consistently outperforms state-of-the-art token compression baselines. 
Notably, at a 25\% retention ratio, \method achieves an average score of 58.9, surpassing the uncompressed Vanilla model (57.3). 
This indicates that learned compression alleviates context rot under token overload, and can even outperform the uncompressed model by removing redundant visual tokens.
\method also delivers substantial efficiency gains: at 10\% retention, it reduces LLM prefill latency by over $16\times$ while preserving 99.5\% of the original accuracy, highlighting its practical value for deploying video MLLMs in latency-sensitive scenarios.

Our key contributions are summarized as follows:
\begin{itemize}
    \item We introduce \method, an RL-based token-level visual compressor for video MLLMs that learns per-token retention 
    policies under token budget constraints by directly optimizing downstream generation quality.
    \item We propose a surprise-augmented state representation to explicitly capture temporal dynamics. 
    To enable effective policy learning, we further introduce a \textit{comprehensive training strategy} that synergizes group-wise policy gradients, a split-advantage estimator, and a progressive pseudo-to-real curriculum.
    \item Extensive experiments show that \method consistently optimizes the accuracy--efficiency trade-off, delivering substantial inference speedups while maintaining competitive performance, and in some settings even surpassing the uncompressed model.
\end{itemize}
\section{Related Works}\label{sec:related_works}

\subsection{Video token compression}

Long-form video understanding with MLLMs is fundamentally constrained by the massive number of visual tokens produced by frame-wise encoding, which is often orders of magnitude larger than in static-image settings. This scale incurs prohibitive computation and exacerbates \emph{context rot}.
Existing video token compression approaches can be broadly grouped by their operational principles~\cite{shao2025tokens}.
\emph{Transformation-based} methods~\cite{zhang2024video,xu2024pllava,maaz2024video,cheng2024videollama} apply uniform downsampling, such as pooling or convolution, across spatial and temporal dimensions.
While simple and efficient, such content-agnostic transformations often remove semantically critical information together with redundant tokens.
\emph{Heuristic} compression methods~\cite{bolya2022token,chen2024image,lin2025boosting,jin2024chat,tao2025dycoke,shen2025fastvid,shao2025holitom,shang2025llava,yang2025visionzip,zhang2025beyond,liu2025video} shorten visual sequences by exploiting either temporal redundancy (e.g., clustering/merging similar frames or tokens) or proxy saliency signals (e.g., pruning low-attention tokens based on attention to a global token such as \texttt{[CLS]}).
While effective in practice, these strategies are typically hand-designed and often decoupled from the model's forward optimization, leaving an optimization gap and failing to explicitly optimize the end-to-end accuracy--compression trade-off.

\subsection{Reinforcement learning for dynamic computation}

Reinforcement learning~\cite{schulman2017proximal,rafailov2023direct,shao2024deepseekmath,yu2025dapo,zheng2025group} has recently achieved immense success and widespread adoption across the domain of Multimodal Large Language Models (MLLMs)~\cite{tao2025dig,wu2025visual,xing2025caprl,zeng2025agentic,liu2025adpo}. It has proven particularly valuable for optimizing dynamic computation scenarios involving non-differentiable and discrete decision-making processes.
In video understanding, Tang \etal~\cite{tang2025tspo} propose TSPO, which formulates keyframe selection as an RL problem for long-form video language understanding.
For architecture-level adaptation, Yue \etal~\cite{yue2024ada} introduce Ada-K routing, using proximal policy optimization to dynamically select the number of activated experts per token in Mixture of Experts models, balancing computational cost and performance.
In reasoning efficiency, Zhang \etal~\cite{zhang2025adaptthink} develops AdaptThink, which uses RL to choose between deeper deliberation and direct generation based on problem difficulty.
These works demonstrate that RL can learn content-aware policies that explicitly trade computation for task performance, providing a natural foundation for our token compression approach.

\section{Preliminaries and Problem Formulation}
\label{sec:preliminaries}

Given an input video $\mathcal{V}$ and a text query $\mathbf{Q}$, a Multimodal Large Language Model first encodes $\mathcal{V}$ into a sequence of visual tokens $\mathbf{X} \in \mathbb{R}^{T \times N \times D}$, where $T$ denotes the number of frames, $N$ the number of tokens per frame, and $D$ the hidden dimension.
The visual tokens are concatenated with the query tokens $\mathbf{Q}$ to form the input to a frozen LLM, which generates an output response $\mathbf{Y}$ autoregressively.
The primary computational bottleneck arises from self-attention in the LLM, whose time and memory complexity scale quadratically with the total sequence length, namely $\mathcal{O}\big((TN + |\mathbf{Q}| + |\mathbf{Y}|)^2\big)$.
For long videos, the visual token count $TN$ dominates the input length, leading to prohibitive inference cost.
This inefficiency limits the practicality of video MLLMs in latency-sensitive and real-time settings.

To alleviate this bottleneck, we aim to compress the visual token sequence $\mathbf{X}$ into a much smaller subset $\mathbf{X}_{\mathrm{comp}} \subseteq \mathbf{X}$, reducing the number of tokens processed by the LLM while preserving model capability.
We formalize this goal as the following constrained optimization problem:
\begin{equation}
    \begin{aligned}
        \min_{\mathbf{X}_{\mathrm{comp}}\subseteq \mathbf{X}}\;& |\mathbf{X}_{\mathrm{comp}}| \\
        \text{s.t.}\; \mathcal{A}(\mathbf{X}_{\mathrm{comp}})&\ge \mathcal{A}(\mathbf{X})-\delta ,
    \end{aligned}
\end{equation}
where $|\mathbf{X}_{\mathrm{comp}}|$ denotes the number of retained visual tokens, $\mathcal{A}(\cdot)$ is a model capability metric, and $\delta \geq 0$ is a user-specified tolerance on performance degradation.
Directly solving this problem is intractable due to the exponential subset search space and the non-differentiability of $\mathcal{A}(\cdot)$ with respect to discrete token selection.
These challenges motivate our reinforcement learning formulation, in which a lightweight policy network learns to select informative tokens adaptively under an explicit token budget.


\section{Methodology}
\label{sec:methodology}

\subsection{Overview}
\label{sec:overview}

\cref{fig:method_overview} provides an overview of \method.
A frozen vision encoder and projector map a long video into visual tokens $\mathbf{X}[t]\in\mathbb{R}^{N\times D}$, and \method inserts a lightweight, trainable token-level compressor before a frozen LLM to dynamically select informative tokens.
To expose temporal dynamics, the compressor conditions on a surprise-augmented state $\mathbf{H}[t]=[\mathbf{X}[t];\Delta\mathbf{X}[t]]$, where $\Delta\mathbf{X}[t]=\mathbf{X}[t]-\mathbf{X}[t-1]$.
A small MLP outputs Bernoulli retention probabilities $p_{t,i}$, defining a stochastic pruning policy over the $T\times N$ tokens.
We optimize the policy with on-policy reinforcement learning using group rollouts. 
For each video, we sample $K$ masks, evaluate each compressed input by the frozen LLM via token-level cross-entropy, and construct an accuracy--sparsity reward with a split advantage for stability.
Training follows a two-stage curriculum from pseudo-videos to real videos.
At inference, we apply a deterministic global top-$K$ rule to retain exactly $\lfloor \rho TN \rfloor$ tokens, where $\rho\in(0,1]$ denotes the user-specified retention ratio, while preserving their original spatiotemporal positions.

\begingroup
\begin{figure*}[t]
    \centering
    \includegraphics[width=\textwidth]{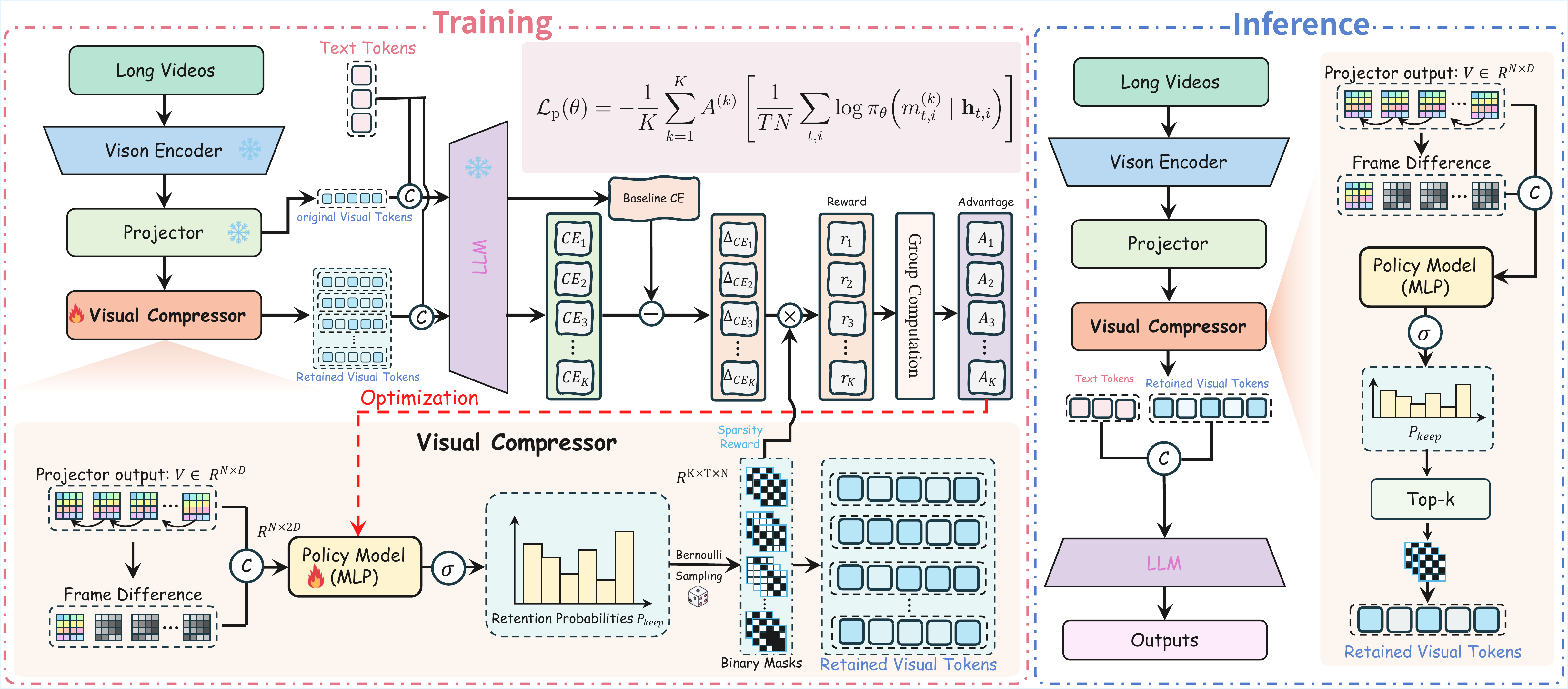}
    \caption{\textbf{\method pipeline.} A lightweight visual compressor 
    is inserted between the frozen vision encoder and the frozen LLM. During training, we optimize a surprise-augmented Bernoulli gating policy with group rollouts and an accuracy--sparsity reward. During inference, we use deterministic global top-$K$ selection to meet a target budget.}
    \label{fig:method_overview}
\end{figure*}
\endgroup

\subsection{Token-Level Visual Compressor}
\label{sec:compressor}

\textbf{Surprise-Augmented State Encoding.}
Identifying informative visual tokens in videos requires modeling both static content and temporal dynamics. However, token embeddings produced by a vision encoder are often highly correlated across adjacent frames due to temporal continuity, yielding redundant representations.
If the policy network is conditioned only on the original embeddings $\mathbf{X}[t]$, it tends to assign similar retention probabilities to tokens in consecutive frames, which hinders the discovery of motion- or change-salient regions.
To make temporal changes explicit and break this symmetry, we augment each token with a \emph{surprise} signal computed from inter-frame residuals:
\begin{equation}
    \Delta\mathbf{X}[t] = \mathbf{X}[t] - \mathbf{X}[t-1] \in \mathbb{R}^{N \times D}, \quad t = 1, \dots, T,
\end{equation}
where we define a virtual zero frame $\mathbf{X}[0]=\mathbf{0} \in \mathbb{R}^{N \times D}$ so that $\Delta\mathbf{X}[1]=\mathbf{X}[1]$.
This residual serves as a temporal high-pass component, emphasizing regions with motion or semantic change while suppressing static redundancy.
We then form the per-frame token representation by concatenating the original embedding and its surprise signal:
\begin{equation}
    \mathbf{H}[t] = [\mathbf{X}[t]; \Delta\mathbf{X}[t]] \in \mathbb{R}^{N \times 2D}.
\end{equation}
The resulting representation provides the policy with explicit cues about temporal variation, enabling content- and motion-aware token selection.

\noindent\textbf{Bernoulli Gating Policy Network.}
The policy network maps each token representation $\mathbf{h}_{t,i} \in \mathbb{R}^{2D}$ through a lightweight multi-layer perceptron (MLP) followed by a sigmoid activation.
The retention probability is computed as:
\begin{equation}
    p_{t,i} = \sigma(\text{MLP}(\mathbf{h}_{t,i})) = \frac{1}{1 + \exp(-\text{MLP}(\mathbf{h}_{t,i}))},
\end{equation}
where $\sigma(\cdot)$ denotes the sigmoid function.
The probabilities $\{p_{t,i}\}$ parameterize independent Bernoulli decisions over tokens, defining the policy $\pi_\theta$.

\subsection{Policy Gradient Optimization}
\label{sec:optimization}

\textbf{Group Rollouts.}
Our visual token compressor induces a stochastic policy $\pi_\theta$ that outputs retention probabilities $p_{t,i}$ for each token.
A discrete compression mask $\mathbf{M}\in\{0,1\}^{T\times N}$ is obtained by sampling
$m_{t,i} \sim \mathrm{Bernoulli}(p_{t,i})$.
The resulting compressed token sequence is $\hat{\mathbf{X}} = \mathbf{X}\odot \mathbf{M}$, where $\odot$ denotes elementwise masking.
Since the sampling operation is non differentiable, we optimize $\pi_\theta$ with policy gradient methods by maximizing the expected reward
$\mathbb{E}_{\mathbf{M}\sim \pi_\theta}\!\left[R(\mathbf{M})\right]$,
where $R(\mathbf{M})$ evaluates the model behavior under the compressed input.
Estimating gradients from a single sampled mask can have high variance.
To stabilize training, we adopt group rollouts following Shao \etal~\cite{shao2024deepseekmath}: for each video, we sample $K$ independent masks $\mathbf{M}^{(1)},\dots,\mathbf{M}^{(K)}$ from the current policy, producing compressed sequences $\hat{\mathbf{X}}^{(k)}=\mathbf{X}\odot \mathbf{M}^{(k)}$. 
The set $\{\hat{\mathbf{X}}^{(k)}\}_{k=1}^K$, together with the original $\mathbf{X}$, is then used to construct a reward signal for policy optimization.

\noindent\textbf{Accuracy and Sparsity Reward.}
To operationalize the objective in \cref{sec:preliminaries}, we design a reward that trades off capability and token budget.
We use the token-level cross-entropy (CE) of the target response under the frozen LLM as a surrogate for model capability.
For rollout $k$, let $\mathrm{CE}^{(k)}$ denote the CE under the compressed tokens and $\mathrm{CE}_{\mathrm{base}}$ the CE under the full tokens.
We define
\begin{align}
    R_{\mathrm{perf}}^{(k)} &= \Delta\mathrm{CE}^{(k)} = \tau \cdot \mathrm{CE}_{\mathrm{base}} - \mathrm{CE}^{(k)}, \\
    R_{\mathrm{comp}}^{(k)} &= S^{(k)} = 1 - \frac{1}{TN}\sum_{t,i} m_{t,i}^{(k)},
\end{align}
where $\tau \geq 1$ controls the allowable performance drop.
A positive $R_{\mathrm{perf}}^{(k)}$ indicates that the compressed input preserves sufficient information, while $R_{\mathrm{comp}}^{(k)}$ directly measures sparsity (i.e., the compression ratio).

\begin{figure}[h]
    \centering
    \includegraphics[width=0.7\linewidth]{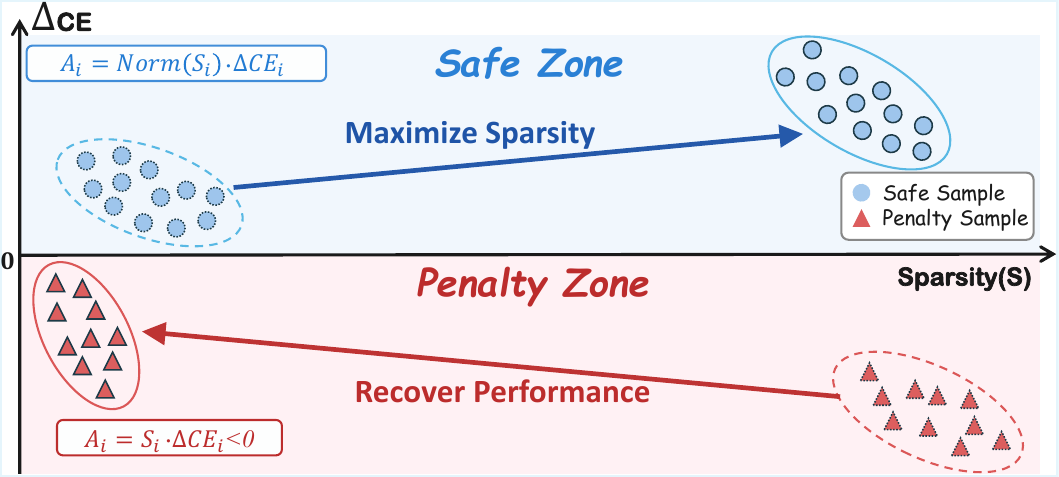}
    \caption{\textbf{Asymmetric advantage for group rollouts.} We separate rollouts into a safe zone ($\Delta \mathrm{CE}>0$) and a penalty zone ($\Delta \mathrm{CE}\le 0$). Successful rollouts are ranked by sparsity via normalized advantages, while violating rollouts receive sparsity-weighted penalties to recover performance.}
    \label{fig:reward}
\end{figure}

\noindent\textbf{Advantage Computation.}
We stabilize policy optimization by computing separate advantages for feasible (constraint-satisfying) and infeasible (constraint-violating) rollouts.
As illustrated in \cref{fig:reward}, we partition the $K$ rollouts into a \emph{safe zone} ($\Delta\mathrm{CE}^{(k)} > 0$) and a \emph{penalty zone} ($\Delta\mathrm{CE}^{(k)} \le 0$).
For safe-zone rollouts, we encourage higher sparsity \emph{among successful rollouts} by using a relative, sparsity-normalized advantage.
For penalty-zone rollouts, we impose a sparsity-weighted negative advantage that increases with the performance violation, pushing the policy to retain more tokens and recover performance. Concretely,
\begin{equation}
    A^{(k)} =
        \begin{cases}
            \Delta\mathrm{CE}^{(k)} \cdot \dfrac{S^{(k)} - \mu_S^+}{\sigma_S^+ + \epsilon}, & \text{if } \Delta\mathrm{CE}^{(k)} > 0, \\[6pt]
            \Delta\mathrm{CE}^{(k)} \cdot S^{(k)}, & \text{if } \Delta\mathrm{CE}^{(k)} \le 0,
        \end{cases}
\end{equation}
where $\mu_S^+$ and $\sigma_S^+$ are the mean and standard deviation of $\{S^{(k)}:\Delta\mathrm{CE}^{(k)} > 0\}$, and $\epsilon$ is a small constant for numerical stability.

\noindent\textbf{Policy Update.}
We optimize the compressor parameters $\theta$ with an on-policy policy-gradient objective. To make the loss scale-invariant to the video length, we normalize by the number of visual tokens:
\begin{equation}
\mathcal{L}_{\mathrm{p}}(\theta) = -\frac{1}{K}\sum_{k=1}^K A^{(k)} \left[ \frac{1}{TN}\sum_{t=1}^T\sum_{i=1}^N \log \pi_\theta\!\left(m_{t,i}^{(k)} \mid \mathbf{h}_{t,i}\right) \right].
\end{equation}
where $\pi_\theta(\cdot \mid \mathbf{h}_{t,i})$ is a Bernoulli policy with parameter $p_{t,i}$. For each token,
\begin{equation}
\log \pi_\theta\!\left(m_{t,i}^{(k)} \mid \mathbf{h}_{t,i}\right)= m_{t,i}^{(k)} \log p_{t,i} + \bigl(1-m_{t,i}^{(k)}\bigr)\log\!\bigl(1-p_{t,i}\bigr).
\end{equation}

\subsection{Curriculum Learning Strategy}\label{sec:curriculum}

Directly optimizing the compressor on long videos is challenging due to the exponential action space and the subtle, noisy nature of real-world motion.
To improve learnability, we adopt a two-stage curriculum that gradually increases the difficulty of temporal dynamics.

\begingroup
\begin{figure*}[t]
    \centering
    \includegraphics[width=1.0\linewidth]{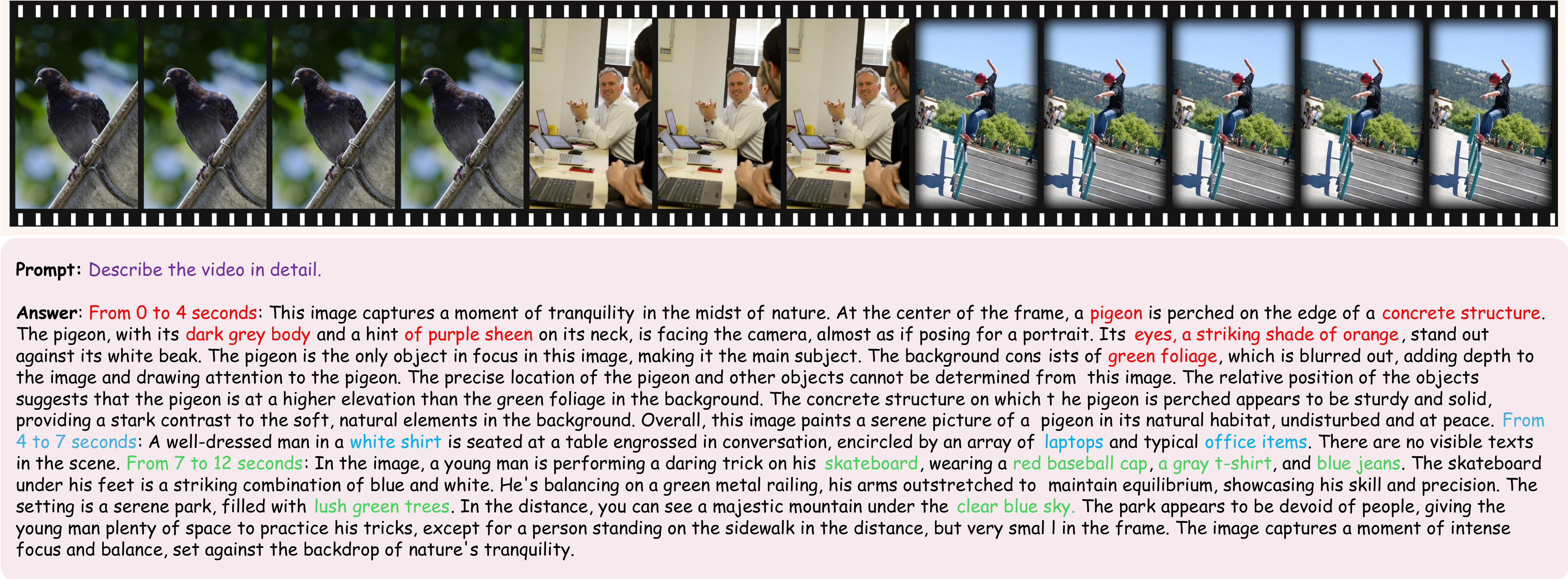}
    \caption{\textbf{Pseudo-video samples for curriculum warm-up.} We construct short clips by repeating each sampled image for several frames and concatenate their captions into a single training target. This synthesis produces near-zero inter-frame residuals within repeated segments and sharp changes at image boundaries, offering high-contrast temporal signals for learning redundancy-aware token pruning.}
    \label{fig:pseudo_video}
\end{figure*}
\endgroup

\noindent\textbf{Warm-up with Pseudo-Videos.}
We first pretrain the compressor on pseudo-videos synthesized from image-caption datasets (\cref{fig:pseudo_video}).
Each pseudo-video is formed by sampling 2--4 images and repeating each image for 3--6 frames to create a short clip;
the corresponding captions are concatenated as the text input. We generate $\sim$10K such synthetic samples.
This construction yields a high-contrast temporal signal.
Unlike real videos, where inter-frame changes are continuous and often dominated by noise, pseudo-videos exhibit an almost binary pattern: consecutive frames are either identical due to repetition (yielding near-zero $\Delta\mathbf{X}$) or change abruptly at image boundaries (yielding large $\Delta\mathbf{X}$).
This discrete structure simplifies credit assignment and enables the policy to quickly learn the association between temporal residuals and token informativeness, i.e., pruning redundant tokens with negligible $\Delta\mathbf{X}$ while retaining tokens around transitions.

\noindent\textbf{Real-Video traninig.}
We then train the compressor on real video-caption datasets containing $\sim$57K pairs.
Equipped with a noise-free prior on how temporal residuals relate to redundancy, the policy can better cope with subtle motion and cluttered backgrounds, refining its selection strategy to separate meaningful dynamics from spurious fluctuations and generalizing the residual-based heuristic to continuous video domains.
This stage reduces the domain gap and stabilizes token selection on in-the-wild videos, improving downstream robustness.

\subsection{Inference Strategy}\label{sec:infer}
At inference time, we use deterministic token selection. Given retention probabilities $p_{t,i}$, we apply a global top-$K$ rule over all $T\times N$ tokens and keep the highest $\lfloor \rho TN \rfloor$ scores, where $\rho$ is the target retention ratio. We keep the original positional indices of the selected tokens, preserving spatial-temporal ordering without changing the positional encoding.

\section{Experiments}
\label{sec:experiments}

\begin{table*}[t]
\centering
\small
\caption{
Main results on Qwen2.5-VL~\cite{bai2025qwen2} across three video understanding benchmarks.
We compare \textbf{\method} with the Vanilla baseline and state-of-the-art token reduction methods at varying retention ratios ($R$).
The best results in each block are marked in \textbf{bold}.
Rows highlighted in \colorbox{greenhighlight}{green} denote our method.
}
\label{tab:main_results}
\resizebox{\textwidth}{!}{
\begin{tabular}{l| c c | c c c c c c | c c}
\toprule
 & & & & & \multicolumn{4}{c|}{VideoMME} & & \\ 
\cmidrule(lr){6-9}
\multirow{-2}{*}{Method} & 
\multirow{-2}{*}{\makecell{Retention \\ Ratio $R$}} & 
\multirow{-2}{*}{\makecell{Max \\ Frames}} & 
\multirow{-2}{*}{LVBench} & 
\multirow{-2}{*}{MLVU} & 
Overall & Short & Medium & Long & 
\multirow{-2}{*}{\makecell{Avg \\ Score}} & 
\multirow{-2}{*}{Acc. (\%)} \\

\rowcolor{gray!40}
Duration & & & 1$\sim$120min & 3$\sim$120min &1$\sim$60min &1$\sim$3min & 3$\sim$30min&30$\sim$60min & & \\
\midrule
\rowcolor{gray!20}
Vanilla
& 100\% & 128 & 38.3 & 69.7  & 63.8 & 74.4 & 64.2 & 52.8 & 57.3 & 100.0 \\
\midrule
DyCoke\textsubscript{\textit{(CVPR'25)}}
& 40\% & 128& 39.8 & 70.0 & 64.9 & 74.3 & \textbf{65.3} & 55.0 & 58.2 & 101.7 \\

HoliTom\textsubscript{\textit{(NeurIPS'25)}}
& 40\% & 128& 39.1 & 69.9 & 61.6 & 73.1 & 59.9 & 51.8 & 56.9 & 99.3 \\

VidCom$^2$\textsubscript{\textit{(EMNLP'25)}}
& 40\% & 128 & 40.2 & \textbf{70.6} & 64.7 & 74.8 & 64.7 & 54.6 & 58.5 & 102.1 \\

FastVID\textsubscript{\textit{(NeurIPS'25)}}
& 40\% & 128& 39.9 & 68.8 & 63.7 & 73.9 & 62.8 & 54.4 & 57.5 & 100.3 \\

\rowcolor{greenhighlight}
\textbf{\method}
& 40\%  & 128& \textbf{40.6} & \textbf{70.6} & \textbf{65.0} & \textbf{75.0} & 65.0 & \textbf{55.1} & \textbf{58.7} & \textbf{102.4} \\

\midrule
DyCoke\textsubscript{\textit{(CVPR'25)}}
& 25\%  & 128& 37.3 & 64.6 & 61.0 & 71.2 & 60.0 & 51.7& 54.3 & 94.8\\

HoliTom\textsubscript{\textit{(NeurIPS'25)}}
& 25\% & 128& 38.2 & 68.4 & 60.3 & 70.2 & 60.1 & 50.4 & 55.6 & 97.1\\

VidCom$^2$\textsubscript{\textit{(EMNLP'25)}}
& 25\% & 128 & \textbf{41.3} & 68.7 & 64.6 & 73.3 & 65.2 & 55.3 & 58.2 & 101.6 \\  

FastVID\textsubscript{\textit{(NeurIPS'25)}}
& 25\% & 128& 39.7 & 68.4 & 63.3 & 74.3 & 61.7 & 53.9 & 57.1 & 99.8 \\

\rowcolor{greenhighlight}
\textbf{\method}
& 25\% & 128&  41.2 & \textbf{70.1} & \textbf{65.3} & \textbf{74.8} & \textbf{65.4} & \textbf{55.7} & \textbf{58.9} & \textbf{102.8}\\

\midrule

HoliTom\textsubscript{\textit{(NeurIPS'25)}}
& 10\% & 128&  36.7 & 65.2 & 56.6 & 66.1 & 55.1 & 48.4 & 52.8 & 92.3 \\

VidCom$^2$\textsubscript{\textit{(EMNLP'25)}}
& 10\% & 128 & 38.6 & 63.4 & 61.8 & 70.4 & 62.4 & 52.6 & 54.6 & 95.3 \\

FastVID\textsubscript{\textit{(NeurIPS'25)}}
& 10\% & 128& 37.8 & 65.0 & 62.0 & 72.9 & 61.7 & 51.4 & 54.9 & 95.9\\

\rowcolor{greenhighlight}
\textbf{\method}
& 10\%  & 128& \textbf{39.6} & \textbf{67.7} & \textbf{63.6} & \textbf{73.6} & \textbf{63.1} & \textbf{53.9} & \textbf{57.0} & \textbf{99.5} \\

\bottomrule
\end{tabular}
}
\end{table*}

\subsection{Experimental Setup}

\noindent\textbf{Models and Benchmarks.}
We evaluate \method on two representative MLLMs to demonstrate generality: \textbf{Qwen2.5-VL}~\cite{bai2025qwen2} and \textbf{LLaVA-Video}~\cite{zhang2024video}.
We report results on three standard video-language benchmarks: \textbf{Video-MME} (w/o subtitles)~\cite{fu2025video}, \textbf{MLVU}~\cite{zhou2025mlvu}, and \textbf{LVBench}~\cite{wang2025lvbench}.
These datasets cover diverse video lengths and question types, enabling a robust evaluation under strict token budgets.

\noindent\textbf{Baselines.}
We compare against \textbf{Vanilla} inference (full visual tokens) and 4 video token compression baselines: \textbf{DyCoke}~\cite{tao2025dycoke}, \textbf{HoliTom}~\cite{shao2025holitom}, \textbf{VidCom$^2$}~\cite{liu2025video} and \textbf{FastVID}~\cite{shen2025fastvid}.
For a fair comparison to our pre-LLM compressor, we evaluate only their \emph{outer-LLM} token pruning/merging components and exclude any \emph{inner-LLM} KV-cache optimizations.
\textbf{DyCoke}~\cite{tao2025dycoke} reduces temporal redundancy via cross-frame token merging but requires a minimum retention rate of $\sim$25\%, so we omit it at the 10\% setting.
\textbf{HoliTom}~\cite{shao2025holitom} performs redundancy-aware temporal segmentation, and we implement its spatiotemporal merging module.
\textbf{VidCom$^2$}~\cite{liu2025video} adaptively adjusts frame-wise compression intensity based on quantified frame uniqueness and selectively preserves distinctive visual tokens both locally and globally.
\textbf{FastVID}~\cite{shen2025fastvid} applies dynamic density-based pruning over temporally ordered segments to preserve essential context.

\noindent\textbf{Training and Evaluation.}
We train the \method policy network with the two-stage curriculum in \cref{sec:curriculum}.
The warm-up stage uses $\sim$10K pseudo-videos built from \textbf{LLaVA-OneVision-Data}~\cite{li2024llava}.
The main stage uses $\sim$57K real video--caption pairs from \textbf{Koala-36M}~\cite{wang2025koala} and \textbf{LLaVA-Video-178K}~\cite{zhang2024video}.
To obtain reliable reward supervision, we filter noisy web text and re-annotate all training samples with strong teacher MLLMs (Qwen3-VL-235B-A22B and Qwen2.5-VL-72B), producing high-quality dense captions that stabilize RL training.
For group-wise optimization, we sample $K=16$ independent masks for each video from the current policy.
We set the performance-tolerance coefficient $\tau$ to 1.01 in warm-up and 1.02 in the main stage.
We use a learning rate of 0.02, and the number of frames is 6--24 in warm-up and up to 128 in the main stage.
During inference, we report results under three retention ratios $\rho\in\{0.10, 0.25, 0.40\}$ to study the accuracy--efficiency trade-off.
More training and evaluation details are provided in Appendix A.

\subsection{Main Results}
\label{sec:main_results}

\noindent\textbf{Results on Qwen2.5-VL.}
\cref{tab:main_results} shows that \method consistently outperforms all baselines on LVBench, MLVU, and Video-MME across retention ratios.
With 40\% retention, \method reaches an average score of 58.7, improving over the strongest baseline (VidCom$^2$~\cite{liu2025video}, 58.5) and even surpassing the uncompressed \textbf{Vanilla} model (57.3).
Under more aggressive compression (25\% retention), \method not only preserves accuracy but slightly improves it, achieving 58.9 on average, demonstrating that the learned policy effectively removes redundant tokens while retaining (and sometimes enhancing) task-relevant information for video understanding.
This advantage is consistent across Video-MME durations (Short/Medium/Long), as well as on LVBench, which emphasizes long-horizon reasoning, and the multilingual MLVU benchmark.
Even at 10\% retention, \method retains 99.5\% of the Vanilla performance and remains clearly ahead of all training-free baselines.

\begin{table}[h]
\centering
\small
\setlength{\tabcolsep}{4pt} 
\caption{
Generalization analysis on Video-MME~\cite{fu2025video} with the LLaVA-Video~\cite{zhang2024video} architecture.
}
\label{tab:llava_video}
\resizebox{0.75\linewidth}{!}{
\begin{tabular}{l| c c | c c c | c c}
\toprule
{Method} & 
\makecell{Retention \\ Ratio $R$} & 
\makecell{Max \\ Frames} & 
 Short & Medium & Long & Overall &
Acc. (\%) \\
\midrule
\rowcolor{gray!20}
Vanilla
& 100\% & 64 & 73.4 & 58.9 & 49.9 & 60.7 & 100  \\
\midrule

FastVID
& 40\% & 64 & 72.7 & 59.2 & 48.4 & 60.1 & 99.0  \\

\rowcolor{greenhighlight}
\textbf{\method}
& 40\% & 64 & 73.2 & 58.0 & 49.7 & \textbf{60.3} & \textbf{99.3} \\

\midrule

FastVID
& 25\% & 64 & 71.9 & 58.1 & 47.8 & 59.3 & 97.7  \\

\rowcolor{greenhighlight}
\textbf{\method}
& 25\% & 64 & 72.6 & 57.8 & 48.8 & \textbf{59.7} & \textbf{98.4} \\

\midrule

FastVID
& 10\% & 64 & 68.3 & 55.1 & 45.8 & 56.4 & 92.9  \\

\rowcolor{greenhighlight}
\textbf{\method}
& 10\% & 64 & 70.0 & 55.3 & 47.3 & \textbf{57.6} & \textbf{94.9} \\

\bottomrule
\end{tabular}
}
\end{table}

\noindent\textbf{Generalization to LLaVA-Video.}
To assess architectural generalization, we further evaluate \method on LLaVA-Video.
As reported in \cref{tab:llava_video}, \method consistently outperforms the strong training-free baseline FastVID across retention ratios on Video-MME.
With 40\% retention, \method attains an overall score of 60.3, exceeding FastVID by 0.2 and preserving 99.3\% of the uncompressed Vanilla accuracy.
The advantage widens as the token budget tightens: at 10\% retention, \method improves over FastVID by 1.2 points (57.6 vs.\ 56.4) while retaining 94.9\% of the original performance.
These results indicate that \method learns a transferable compression policy rather than overfitting to a particular backbone, and it can be effectively applied to other video MLLMs.

\subsection{Efficiency Analysis}
\label{sec:efficiency}

A key objective of token compression is to reduce inference cost while preserving accuracy.
As summarized in \cref{tab:efficiency}, \method yields substantial savings in both token count and latency with minimal overhead.
These gains arise because reducing visual tokens shortens the LLM input sequence, and the FLOPs of self-attention scale quadratically with sequence length.
Concretely, at a 25\% retention ratio, \method reduces the number of visual tokens from 42{,}566 to 10{,}733 (a 75\% reduction), translating to an approximate $6\times$ speedup in the LLM forward pass.
Under more aggressive compression (10\% retention), the visual tokens drop to 4{,}348, resulting in a $16.2\times$ end-to-end speedup in prefill time.
The additional computation of the lightweight compressor is negligible, contributing less than 1\% of total inference time, effectively shifting computation from the expensive LLM to an efficient policy network.
Importantly, these efficiency gains do not come at the expense of capability.
As shown in the last column of \cref{tab:efficiency}, \method maintains and in some settings slightly improves accuracy relative to the uncompressed model on Video-MME across retention ratios.
Overall, \method provides a favorable accuracy--efficiency trade-off, making video MLLMs more practical for latency-sensitive deployment.

\begin{table}[h]
\centering
\small
\caption{Efficiency Comparison on Qwen2.5VL-7B~\cite{bai2025qwen2} with VideoMME~\cite{fu2025video}. We report the exact number of visual tokens, total tokens (including system/user text), and prefill latency.}
\label{tab:efficiency}
\small
\setlength{\tabcolsep}{3pt}
\resizebox{0.9\linewidth}{!}{
\begin{tabular}{l | c | c c c | c}
\toprule
\multirow{2}{*}{Method} & \multirow{2}{*}{\makecell{\# Tokens \\ (Visual / Total)}} & \multicolumn{3}{c|}{Prefill Time (ms)} & \multirow{2}{*}{Acc.} \\
\cmidrule(lr){3-5}
 & & Compressor & LLM Fwd. & Total & \\
\midrule
\rowcolor{gray!15}
Vanilla & 42566 / 42657 & -- & $\res{8539}{2231}$ & $\res{8539}{2231}$ (1.0$\times$) & 63.8 (100.0\%) \\

\textbf{\method ($R$=40\%)} & \textbf{17117 / 17208} & \textbf{$\res{18.6}{7.5}$} & \textbf{$\res{2441}{583}$} & \textbf{$\res{2473}{587}$ (3.5$\times$)} & \textbf{65.0 (101.9\%)} \\

\textbf{\method ($R$=25\%)} & \textbf{10733 / 10824} & \textbf{$\res{18.6}{8.9}$} & \textbf{$\res{1386}{318}$} & \textbf{$\res{1417}{326}$ (6.0$\times$)} & \textbf{65.3 (102.4\%)} \\

\textbf{\method ($R$=10\%)} & \textbf{4348 / 4439} & \textbf{$\res{18.5}{8.5}$} & \textbf{$\res{504}{107}$} & \textbf{$\res{527}{110}$ (16.2$\times$)} & \textbf{63.6 (99.7\%)} \\

\bottomrule
\end{tabular}
}
\end{table}

\subsection{Ablation Study}
\label{sec:ablation}

\noindent\textbf{Ablation on Curriculum Learning.}
We ablate the two-stage curriculum in \cref{sec:curriculum} to assess its contribution.
\cref{tab:ablation_curriculum_learning} compares a policy trained only on pseudo-videos (\textbf{Pseudo}) against one further adapted on real videos (\textbf{Pseudo + Real}).
Training on pseudo-videos alone already yields a competitive compressor, particularly at higher retention (e.g., 58.5 average at $R=40\%$).
However, additional training on real videos consistently improves performance across benchmarks and retention ratios.
The gains are most pronounced under aggressive compression: at $R=10\%$, the average score increases from 55.7 to 57.0 after real-video adaptation.
These results validate the curriculum design: pseudo-video warm-up provides a strong initialization for content selection, while real-video training is necessary to handle the nuanced and noisy temporal dynamics of real-world videos.

\begin{table}[h]
\centering
\small
\setlength{\tabcolsep}{4pt}
\caption{
Ablation study on the training data composition across different retention ratios $R$.
We compare the performance of the model trained solely on \textbf{Pseudo-Videos} versus the model further trained on \textbf{Real Videos}.
The \textbf{Vanilla} baseline ($R=100\%$) is provided for reference.
}
\label{tab:ablation_curriculum_learning}
\resizebox{0.7\linewidth}{!}{
\begin{tabular}{l|c|c|ccc|c}
\toprule
Method & Training Data & $R$ & LVBench & MLVU & VideoMME & Avg \\
\midrule
\rowcolor{gray!10}
Vanilla & - & 100\% & 38.3 & 69.7 & 63.8 & 57.3 \\

\midrule
\multirow{2}{*}{\textbf{\method}} & Pseudo & \multirow{2}{*}{40\%} & 40.5 & 70.1 & 64.9 & 58.5 \\ %
 & Pseudo + Real & & \textbf{40.6} & \textbf{70.6} & \textbf{65.0} & \textbf{58.7} \\

\midrule
\multirow{2}{*}{\textbf{\method}} & Pseudo & \multirow{2}{*}{25\%} & 40.4 & 69.1 & 65.2 & 58.2 \\ %
 & Pseudo + Real & & \textbf{41.2} & \textbf{70.1} & \textbf{65.3} & \textbf{58.9} \\

\midrule
\multirow{2}{*}{\textbf{\method}} & Pseudo & \multirow{2}{*}{10\%} & 38.3 & 65.7 & 63.2 & 55.7 \\ %
 & Pseudo + Real & & \textbf{39.6} & \textbf{67.7} & \textbf{63.6} & \textbf{57.0} \\

\bottomrule
\end{tabular}
}
\end{table}

\noindent\textbf{Ablation on Surprise-Augmented State Encoding.}
We ablate the proposed surprise-augmented state encoding by removing the inter-frame residual signal and conditioning the policy only on the raw token embeddings $\mathbf{X}[t]$.
As shown in \cref{tab:ablation_surprise}, this variant consistently underperforms the full \method model across all retention ratios on Video-MME, with the gap becoming more pronounced under tighter budgets.
This degradation supports our motivation: due to temporal continuity, $\mathbf{X}[t]$ is often highly correlated across adjacent frames, causing the policy to assign nearly identical retention probabilities over time and making it difficult to identify change-salient regions.
In contrast, the residual $\Delta\mathbf{X}[t]$ acts as a temporal high-pass component that highlights motion or semantic changes while suppressing static redundancy, thereby breaking temporal symmetry and enabling more informed keep/drop decisions.
Overall, the ablation confirms that explicitly encoding temporal ``surprise'' is critical for effective token selection in video understanding.

\begin{table}[h]
\centering
\small
\setlength{\tabcolsep}{5pt}
\caption{
Ablation study on the impact of the \textbf{Surprise-Augmented State Encoding}.
We compare the performance on VideoMME~\cite{fu2025video} (Short, Medium, Long) and the Overall score across different retention ratios.
}
\label{tab:ablation_surprise}
\resizebox{0.7\linewidth}{!}{
\begin{tabular}{l|c|ccc|c}
\toprule
Method & $R$ & Short & Medium & Long & Overall \\
\midrule
\rowcolor{gray!15}
Vanilla & 100\% & 74.4 & 64.2 & 52.8 & 63.8 \\

\midrule

w/o Residual & \multirow{2}{*}{40\%} & 74.4 & \textbf{65.3} & 54.3 & 64.7 \\
\textbf{\method} (w/ Residual) & & \textbf{75.0} & 65.0 & \textbf{55.1} & \textbf{65.0} \\

\midrule

w/o Residual & \multirow{2}{*}{25\%} & 74.8 & 64.9 & 54.8 & 64.5 \\
\textbf{\method} (w/ Residual) & & \textbf{74.8} & \textbf{65.4} & \textbf{55.7} & \textbf{65.3} \\

\midrule

w/o Residual & \multirow{2}{*}{10\%} & 71.4 & 62.0 & 52.6 & 62.0 \\
\textbf{\method} (w/ Residual) & & \textbf{73.6} & \textbf{63.1} & \textbf{53.9} & \textbf{63.6} \\

\bottomrule
\end{tabular}
}
\end{table}

\begingroup
\begin{figure*}[h]
    \centering
    \includegraphics[width=\linewidth]{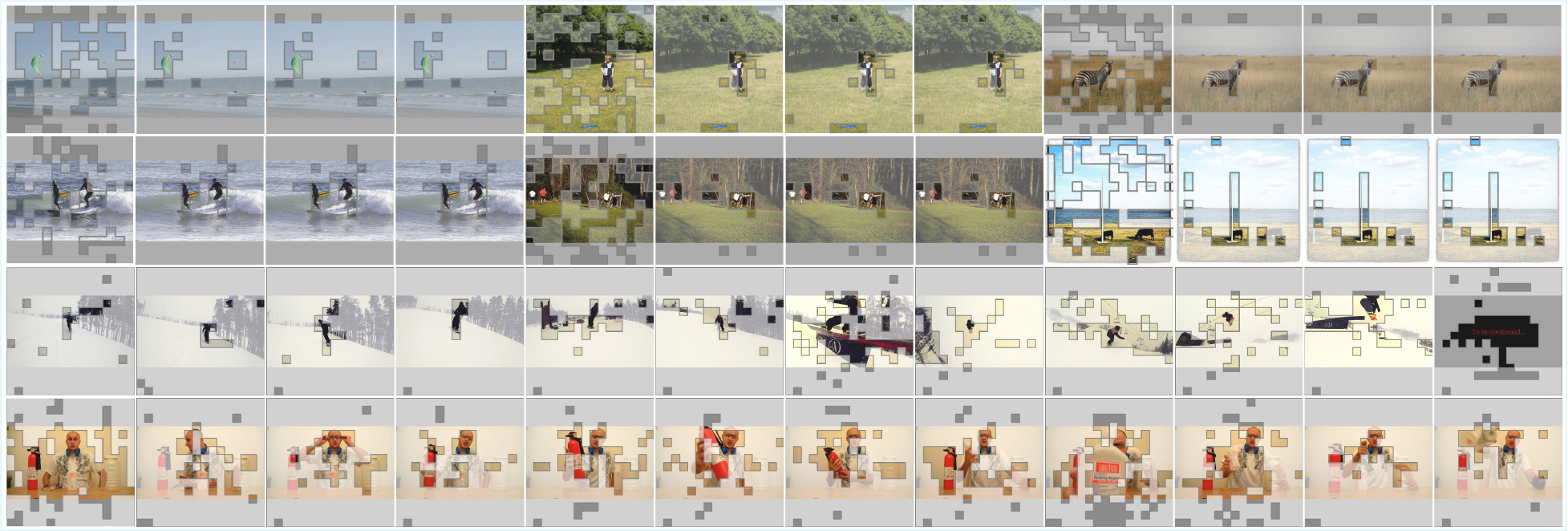}
    \caption{\textbf{Qualitative visualization of \method token masks.} Top two rows: six pseudo-video inputs constructed from static images, where \method concentrates kept tokens on salient objects and action-related regions while pruning static backgrounds. Bottom two rows: two real-video examples, showing temporally adaptive masks that shift as motion and scene content evolve.}
    \label{fig:qualitative_analysis}
\end{figure*}
\endgroup

\subsection{Qualitative Analysis}
\label{sec:qualitative}

\cref{fig:qualitative_analysis} presents a qualitative case study of \method by visualizing the binary token masks produced by the learned compressor.
The visualization reveals two consistent behaviors.
First, \method exhibits strong spatial selectivity: it allocates most retained tokens to primary entities and action related regions, such as humans, animals, and their contact areas with tools or the environment, while pruning large portions of static backgrounds like sky, grass, walls, and water.
Second, \method is temporally adaptive: as motion patterns evolve or the scene content changes, the mask shifts its focus to newly informative regions and reduces allocation to areas that become temporally redundant.
These patterns align with our surprise augmented state encoding, where inter frame residual cues explicitly expose temporal change and enable more decisive keep or drop decisions.
Overall, the masks provide an intuitive explanation for \method's favorable accuracy and efficiency trade-off under strict token budgets.
Additional qualitative examples are included in the supplementary material.

\section{Conclusion}\label{sec:conclusion}

We study the efficiency--accuracy bottleneck of long-context video MLLMs, where the quadratic cost of attention and the accompanying \emph{context rot} jointly hinder reliable video understanding at scale.
To address this challenge, we formulate visual token compression as a reinforcement learning problem and propose \textbf{\method}, a framework centered on a lightweight token-level visual compressor.
\method learns a dynamic compression policy via surprise-augmented state encoding for temporal change awareness, a group-based on-policy optimization scheme with a split advantage objective, and a two-stage curriculum that transitions from pseudo- to real videos.
Together, these components enable content- and dynamics-aware token selection under strict token budgets.
Extensive experiments show that \method establishes a new state of the art across multiple benchmarks, consistently outperforming prior compression methods.
Notably, \method can even surpass the uncompressed baseline at high compression rates (e.g., 25\% retention), suggesting that targeted redundancy removal can mitigate context rot rather than merely preserve performance.
Meanwhile, \method delivers substantial computational benefits, achieving over $16\times$ prefill speedup, highlighting its practicality for latency-sensitive deployment of video MLLMs.


%
%
\bibliographystyle{splncs04}
\bibliography{main}

\end{document}